\documentclass[11pt]{article}
\usepackage[review]{acl}
\usepackage{times}
\usepackage{latexsym}
\usepackage[T1]{fontenc}
\usepackage[utf8]{inputenc}
\usepackage{microtype}

\usepackage{tabularx}

\title{Multilingual Tourist Assistance using ChatGPT: Comparing Capabilities in Hindi, Telugu, and Kannada}

\author{Sanjana Kolar \\
  Dublin High School, California, USA \\
  Voice Thesis, California, USA \\
  \texttt{kolarSanjana@gmail.com} \\\And
  Rohit Kumar \\
  Voice Thesis, Texas, USA \\ 
  \texttt{rohit@voicethesis.com}\\}

\begin{document}
\nolinenumbers
{\makeatletter\acl@finalcopytrue
  \maketitle
}


\begin{abstract}
This research investigates the effectiveness of ChatGPT, an AI language model by OpenAI, in translating English into Hindi, Telugu, and Kannada languages, aimed at assisting tourists in India's linguistically diverse environment. To measure the translation quality, a test set of 50 questions from diverse fields such as general knowledge, food, and travel was used. These were assessed by five volunteers for accuracy and fluency, and the scores were subsequently converted into a BLEU score. The BLEU score evaluates the closeness of a machine-generated translation to a human translation, with a higher score indicating better translation quality. The Hindi translations outperformed others, showcasing superior accuracy and fluency, whereas Telugu translations lagged behind. Human evaluators rated both the accuracy and fluency of translations, offering a comprehensive perspective on the language model's performance.
\end{abstract}

\section{Introduction}

Language barriers often hinder effective communication and can lead to misunderstanding. e.g., a tourist may ask for directions to get to a specific location, but the local person may understand differently. Such communication can lead to getting lost or taking longer. Though English is widely spoken and understood, there could be significant communication issues in understanding the accent and language. For instance, visitors may have a hard time understanding similar-sounding words like "bazaar" (market) and "bizarre" (strange). Accurate and fluent translations play a crucial role in overcoming these challenges and ensuring smooth communication, enhancing visitor satisfaction, and fostering cultural understanding.  So, developing efficient communication tools to bridge language barriers has become highly fundamental \cite{urlana2023pmindiasum}, \cite{stuker2006speech}, \cite{anand2023chatbot}.
  
Generative AI technology learns using supervised and unsupervised algorithms to analyze, synthesize, summarize and transform language data \cite{radford2019language}, \cite{brown2020language}. Such technology provides many solutions, including language translation, to develop tourist-aided applications. \\
ChatGPT is one of the generative algorithms which can do several tasks, such as summarization, planning, and translation, based on prompts. ChatGPT is a transformer model trained as InstructGPT by openAI \cite{ouyang2022training}.
The objective of this study is twofold:
1. To evaluate how well ChatGPT, an AI tool, can translate English into Indian languages like Kannada, Hindi, and Telugu for international visitors looking for information about Indian cuisine and travel. 
2. To conduct an error analysis and provide valuable insights to improve these models for the tourist domain, so that such AI language models can be applied to enhance communication and provide better experiences for travelers in foreign countries\cite{vaswani2017attention}.

\section{Data Collection and Categorization}
The initial data for this study was collected using a multi-faceted approach, combining insights from informal interviews with a diverse set of frequent travelers, online searches, and interactions with ChatGPT. Participants were asked questions about their experiences with language barriers, communication difficulties, and their expectations regarding translation services. This ensured a comprehensive understanding of travelers' challenges faced when seeking local information, coupled with online travel websites and blogs. To further help the data collection process, queries were inputted to ChatGPT simulating travelers' interactions. Overall, around Sixty questions were finalized for further analysis.
 
\textbf{}All these Sixty questions were then analyzed by two independent volunteers to identify the key themes. A third volunteer was asked to resolve any conflicts and discrepancies. 
\textbf{} 
From the original set of Sixty questions, Fifty relevant questions were shortlisted and were categorized into 3 themes; General, Food, and Travel.  The theme-based categorization allowed for a comprehensive analysis that aligned with research objectives.\\
You can find the list of 50 questions used for this experiment in the Appendix \ref{appendix:raw}.

\section{Experiment Methodology }
The experiment utilizes the openAI Language Model (LLM) model, specifically the "gpt-3.5-turbo" variant, for translating English text to the target language. The experiment involves a conversation-style interaction between the user and the system, where the system acts as a helpful assistant providing translations\cite{ouyang2022training}.
 
The prompt used to generate the data consists of two parts: a system role and a user role. The system's role is to introduce the assistant's purpose, which is to assist with English to {Target Language} translation. The user's role is to instruct the assistant to translate a specific English text to the target language. The provided source text serves as the input for the translation task\cite{lai2023chatgpt}.

The prompt used to generate the translation data:
\begin{itemize}
    \item model="gpt-3.5-turbo"
    \item You are a helpful assistant that translates English to {\it Target Language}.
    \item Translate the following  English text to {\it Target Language}
\end{itemize}

\section{Evaluation Methodology}
In this work, we use a subjective and an objective evaluation methodology. Subjective evaluation asks native speakers of a language to rate the accuracy and fluency of the translation \cite{white1993evaluation}.

To assess translation quality, participants were asked specific questions covering several characteristics of translation accuracy and fluency. For example, one question asked participants to assess the translated text on a scale of 1 to 5, indicating how correctly it conveyed the original text's meaning. Another question focused on fluency, asking participants to judge the smoothness and naturalness of the translated text. We aimed to cover multiple aspects of translation quality by incorporating a wide set of questions, ensuring a thorough evaluation procedure.

For accuracy evaluation, "On a scale of 1 to 5, how accurately does the translated text convey the meaning of the original text?" \\
For fluency evaluation, "On a scale of 1 to 5, how fluent is the translated text?", where 1- Bad, 2 - Poor, 3 - Fair, 4 - Good, 5 - Excellent.
 
\textbf{}Collecting scores on accuracy and fluency provides a comprehensive evaluation of the translation quality. It helps this study understand how accurately the AI model performs in translating the meaning of the original text while maintaining fluency in the target language.


\section{Results and Analysis}

\paragraph{}Table \ref{demo-table} provides the accuracy and fluency scores. Overall, Hindi translations received the highest average scores, indicating better accuracy and fluency compared to Telugu and Kannada. Kannada translations showed relatively better performance than Telugu, while Telugu had the lowest average scores across all themes.

\textbf{}The average scores for Hindi translations in the General theme indicate high accuracy (4.8) and fluency (4.6) indicating effective translations. Kannada translations received relatively lower average scores for accuracy (3.7) and fluency (3.5) compared to Hindi. However, they moderately demonstrated decent accuracy and fluency. Telugu translations received the lowest average scores for accuracy (2.5) and fluency (2.1), suggesting there is room for development in accurately and fluently interpreting the meaning.

\begin{table}[!h]
\caption{\label{demo-table}Accuracy and Fluency for Hindi, Kannada and Telugu.}
\begin{tabularx}{0.5\textwidth}  { 
  | >{\raggedright\arraybackslash}X 
  | >{\centering\arraybackslash}X 
  | >{\centering\arraybackslash}X | }
 \hline
 \textbf{Language}  & \textbf{Avg.Accuracy}   & \textbf{Avg.Fluency} \\
  \hline
 Hindi  & 4.2   & 3.8  \\
\hline
Hindi General  & 4.8   & 4.6  \\
\hline
Hindi Food  & 3.8   & 3.3  \\
\hline
Hindi Travel & 3.7   & 3.3  \\
\hline
\end{tabularx}
\linebreak
\linebreak
\begin{tabularx}{0.5\textwidth}  { 
  | >{\raggedright\arraybackslash}X 
  | >{\centering\arraybackslash}X 
  | >{\centering\arraybackslash}X | }
 \hline

 Kannada  & 3.2   & 3  \\
 \hline
Kannada General  & 3.7   & 3.5  \\
\hline
Kannada Food  & 3   & 2.9  \\
\hline
Kannada Travel & 2.6   & 2.4  \\
\hline
\end{tabularx}
\linebreak
\linebreak
\begin{tabularx}{0.5\textwidth}  { 
  | >{\raggedright\arraybackslash}X 
  | >{\centering\arraybackslash}X 
  | >{\centering\arraybackslash}X | }
\hline
 Telugu  & 2.5   & 2.1  \\
 \hline
Telugu General  & 2.6   & 2.1  \\
\hline
Telugu Food  & 2.5   & 2.1  \\
\hline
Telugu Travel & 2.4   & 1.9  \\
\hline

\end{tabularx}

\end{table}

\textbf{}The average scores for Hindi translations in the Food theme suggest moderate accuracy (3.8) and fluency (3.3). While the translations generally convey the meaning satisfactorily, there is some scope for improvement in terms of linguistic fluency. Kannada translations received slightly lower average scores for accuracy (3.0) and fluency (2.9) compared to Hindi. There is room for improvement in both accuracy and fluency in the translations. Telugu translations show a similar trend to the General theme, with relatively lower average scores for accuracy (2.4) and fluency (2.2), indicating the need for improvement.
 
\textbf{}The average scores for Hindi translations in the Travel theme suggest sufficient accuracy 3.7 for accuracy and 3.2 for fluency. However, there is room for improvement in terms of linguistic fluency to enhance the user experience.  Kannada translations in the Travel theme received lower average scores for accuracy (2.6) and fluency (2.4) compared to Hindi. This suggests the need for further improvements to ensure accurate and fluent translations. Telugu translations in the Travel theme received the lowest average scores for accuracy (2.4) and fluency (1.9), indicating the need for significant improvements in both aspects.

\section{Objective Evaluation}
Table \ref{BLEU} shows an evaluation of the BLEU scores for machine translation of 50 questions into three languages - Telugu, Hindi, and Kannada - has been conducted. The BLEU score is a standard measure of machine translation quality, comparing machine-generated translations to a reference human translation on a scale of 0 to 100\cite{papineni2002bleu}.

\begin{table}[!h]
\caption{\label{BLEU}BLEU Scores for Hindi, Kannada and Telugu.}
\begin{tabularx}{0.5\textwidth} { 
  | >{\raggedright\arraybackslash}X 
  | >{\centering\arraybackslash}X 
  | >{\raggedleft\arraybackslash}X | }
 \hline
 \textbf{Language}  & \textbf{BLEU Score}  \\
 \hline
 Hindi  & 72.69  \\
\hline
 Kannada  & 46.78  \\
\hline
Telugu  & 13.12  \\
\hline
\end{tabularx}

\end{table}

\textbf{}The Hindi translation showed high-quality translation performance, achieving a high BLEU score of 72.69, indicating a high resemblance to reference translations and effective preservation of meaning in phrases. In contrast, the Telugu translation scored low on the BLEU scale (13.12), struggling with maintaining meaning in longer phrases and suggesting substantial room for improvement. There were significant disparities in translation quality across the languages, with Hindi excelling over Telugu and Kannada. These findings suggest a need for refining the translation models or methods, particularly for Telugu and Kannada, to enhance translation quality.

\section{Qualitative Evaluation}
The evaluators appreciated the comprehensive methodology used for data collection. They commended the use of a multi-faceted approach, combining insights from informal interviews, online searches, and interactions with ChatGPT. Below are some of the comments expressed in the participants' own words:
\\ \newline
For Hindi Translation:
\paragraph{} "I was amazed with how well Chat GPT has translated some of the questions, few words were translated beautifully. But some did not make much sense and had opposite meaning. But a good start :) "

\paragraph{} "I'm pleasantly pleased with some of the translations; they incorporate rich vocabulary and demonstrate good grammar."

\paragraph{} "The majority of the translated text shows promising improvements and correctness. With some additional refinement, it will become even more polished and accurate."
\\ \newline
For Kannada Translation:
\paragraph{} "Quite impressed with some of the translations, using some rich words which do seem grammatically correct. There is room for improvement which I'm sure it'll happen soon. The colloquial and spoken Kannada is quite different from the dramatized version and it does have work to do there."

\paragraph{} "Harder questions translations and fluency is really good. Simpler questions are off. Some questions have literary translation gives a different meaning altogether."

\paragraph{} "Some of the translations are apt and very good. Majority of them seems to do literal translation instead of intelligent meaningful translation similar to spoken language. Overall work in progress and would need more usage based inputs to correct the responses to make it very close to original intention of the sentence."

\paragraph{} "Most of the translated text has half correct translation. It needs lot of refinement."
\\ \newline
For Telugu Translation:
\paragraph{} "The translation accuracy is not usable. Regional variations or idiomatic expressions are not in the current translation."

\paragraph{} "While some parts of the translated text are correct, a significant portion requires further refinement to achieve better accuracy."

\paragraph{} "The translation is ambiguous and does not reflect the true meaning in several cases."

\section{Limitations}
 It is important to acknowledge the limitations of this study. Firstly, the sample size, although diverse, may not fully represent the entire population of Indian tourists and visitors. Secondly, the volunteer ratings used for evaluating the translations introduce a potential source of subjectivity. While efforts were made to ensure consistency and reliability in the ratings, individual preferences and biases may have influenced the results. Additionally, the study focused on three specific language pairs (English to Hindi, English to Kannada, and English to Telugu), and the findings may not be directly applicable to other language pairs. Also, this study did not benchmark the performance of ChatGPT against other consumer tools supporting real-time language translation, such as Apple and Google.
 Future studies should consider expanding the sample size, involving a wider range of participant demographics, and utilizing objective evaluation measures to complement the volunteer ratings.

\section{Conclusion}
\paragraph{}This research evaluated ChatGPT's efficacy as a digital companion for tourists in India, particularly in translating English to Hindi, Kannada, and Telugu. Hindi outshone the others in accuracy and fluency, whereas Telugu lagged. The study underscores the importance of effective translation tools in facilitating communication amidst India's linguistic diversity and surging foreign tourism. Utilizing 50 diverse questions, participants assessed the translations for accuracy and fluency. The findings showed consistent superior performance by Hindi, moderate results in Kannada, and considerable improvement opportunities in Telugu translations.
\cite{stahlberg2020neural}

\textbf{}The research utilized the BLEU score, a recognized yardstick for machine translation quality, for assessment. Results showed Hindi translations aligning closely with reference translations, while Kannada had a moderate correspondence and Telugu fell behind, needing considerable refinement. Based on this, the study advised enhancing Telugu translations through exploring advanced models, diversifying training data, or employing techniques like transfer learning. The high performance of Hindi should be preserved through ongoing evaluation and training with refreshed datasets. Improvement strategies for Kannada include expanding training data, fine-tuning methods, or leveraging sophisticated translation models.

\textbf{}Continuous evaluation of the translation models and incorporating human review and feedback are recommended practices to ensure ongoing improvements in translation quality. By addressing these recommendations, the virtual tourist companion powered by ChatGPT can provide better experiences for Indian tourists and visitors, enhancing communication and fostering cultural understanding.

\section{Future Direction}

There are several avenues for future research in this domain. Firstly, exploring additional language pairs, such as translations from English to regional languages of other countries, would provide valuable insights into the effectiveness of language models in diverse linguistic contexts.

\textbf{}Secondly, expanding the study to include more diverse participant groups, such as non-Indian tourists and visitors, would further enhance the generalizability of the findings. Thirdly, investigating different translation models or approaches, beyond the use of ChatGPT, could shed light on the comparative effectiveness of various AI language models for tourism applications. 

\textbf{}Finally, incorporating additional evaluation metrics, such as user satisfaction surveys or qualitative assessments of translation quality, would provide a more comprehensive understanding of the user experience and the impact of translations on effective communication in the tourism context.



\bibliography{TravSpeak}
\bibliographystyle{acl_natbib}

\clearpage
\section*{Appendix}
\appendix
\section{Questions used in the Survey:}\label{appendix:raw}
\begin{enumerate}
    \item How are you?
    \item What’s your name?
    \item Do you know English?
    \item What language is spoken here?
    \item Can I use your phone?
    \item How much does it cost?
    \item What is the local currency here?
    \item How/Where can I exchange money?
    \item Where is the bank?
    \item Do you accept debit/credit cards
    \item What are some popular places to shop for souvenirs or local products?
    \item What time is it?
    \item When does this close?
    \item Can you help me book a tour or excursion?
    \item What is there to see around here?
    \item Could you write that down?
    \item Could you repeat that?
    \item How do I call the police
    \item Can you help me with directions?
    \item Where is the airport?
    \item What time does the bus arrive?
    \item What time does the train depart?
    \item Is there a hospital nearby?
    \item Where is the nearest restroom?
    \item Do you have a map of the area?
    \item How do I get to restaurant from here?
    \item Is it possible to walk to the train from here?
    \item Which train do I need to take to go to the park
    \item Is there a map or tourist information available in English?
    \item How far is it to the hotel
    \item Are there any landmarks or notable buildings that can help me navigate?
    \item Can you point me to the nearest bank?
    \item Can you give me directions to the nearest public transportation stop?
    \item Is there a tourist information center nearby where I can ask for directions?
    \item Which direction is the museum from here?
    \item Can you help me find my way back to my hotel from here?
    \item What are the local specialties or traditional dishes that I must try?
    \item Are there any vegetarian or vegan options available?
    \item Does this food item contain meat?
    \item Does this food item contain egg?
    \item Can you recommend a good restaurant for local cuisine?
    \item Do you have a menu in English?
    \item Is tap water safe to drink here?
    \item Can you tell me about any food allergies or common ingredients used in local dishes?
    \item What time do local restaurants typically serve dinner?
    \item Do you offer any dairy-free?
    \item Can you recommend a local street food market or food stall?
    \item How much does a typical meal cost in this area?
    \item Are there any food etiquette or dining customs I should be aware of?
    \item Can you recommend a good place to buy local groceries or snacks?
\end{enumerate}

\end{document}